\pdfoutput=1

\documentclass[11pt]{article}

\usepackage[]{EMNLP2023}

\usepackage{times}
\usepackage{latexsym}

\usepackage[T1]{fontenc}

\usepackage[utf8]{inputenc}

\usepackage{microtype}

\usepackage{inconsolata}

\usepackage{booktabs}
\usepackage{caption}
\usepackage{color}
\usepackage{xcolor}
\usepackage{graphicx}
\usepackage{latexsym}
\usepackage{makecell}
\usepackage{soul}
\usepackage{subcaption}
\usepackage{times}
\usepackage{amssymb}
\usepackage{amsmath}
\usepackage{amsbsy}
\usepackage{enumitem}
\usepackage{bm}

\title{Rigorously Assessing Natural Language Explanations of Neurons}

\newcommand{\authorspace}{\hspace{1pt}}

\author{Jing Huang$^{1}$  \authorspace 
Atticus Geiger$^{1,2}$  \authorspace
Karel D'Oosterlinck$^{1,3}$  \authorspace
Zhengxuan Wu$^{1}$  \authorspace 
Christopher Potts$^{1}$ \\[1ex]
  ${}^{1}$Stanford University\qquad   ${}^{2} $Pr(Ai)$^2$R Group\qquad ${}^{3}$Ghent University – imec \\[1ex]
  \texttt{\{hij, atticusg, kldooste, wuzhengx, cgpotts\}@stanford.edu} \\
  }

\usepackage{stmaryrd}
\newcommand{\sem}[1]{\ensuremath{\llbracket#1\rrbracket}}
\newcommand{\word}[1]{\emph{#1}}
\newcommand{\GetVals}{\texttt{GetVals}}
\newcommand{\Explain}{\textsc{Explain}}
\newcommand{\CausalExplain}{\textsc{CausalExplain}}
\newcommand{\explanation}[1]{\emph{#1}}

\begin{document}
\maketitle
\begin{abstract}
Natural language is an appealing medium for explaining how large language models process and store information, but evaluating the faithfulness of such explanations is challenging. To help address this, we develop two modes of evaluation for natural language explanations that claim individual neurons represent a concept in a text input. In the \emph{observational mode}, we evaluate
claims that a neuron $a$ activates on all and only input strings that refer to a concept picked out by the proposed explanation $E$. In the \emph{intervention mode}, we construe $E$ as a claim that the neuron $a$ is a causal mediator of the concept denoted by $E$. We apply our framework to the GPT-4-generated explanations of GPT-2~XL neurons of \citet{bills2023language} and show that even the most confident explanations have high error rates and little to no causal efficacy. We close the paper by critically assessing whether natural language is a good choice for explanations and whether neurons are the best level of analysis.
\end{abstract}

\section{Introduction}

The ability to generate natural language explanations of large language models (LLMs) would be an enormous step forward for explainability research. Such explanations could form the basis for safety assessments, bias detection, and model editing, in addition to yielding fundamental insights into how LLMs represent concepts. However, we must be able to verify that these explanations are \emph{faithful} to how the LLM actually reasons and behaves. 

What criteria should we use when assessing the faithfulness of natural language explanations? Without a clear answer to this question, we run the risk of adopting incorrect (but perhaps intuitive and appealing) explanations, which would have a severe negative impact on all the downstream applications mentioned above. 

\begin{figure}
     \centering
     \includegraphics[width=0.75\linewidth,angle=270]{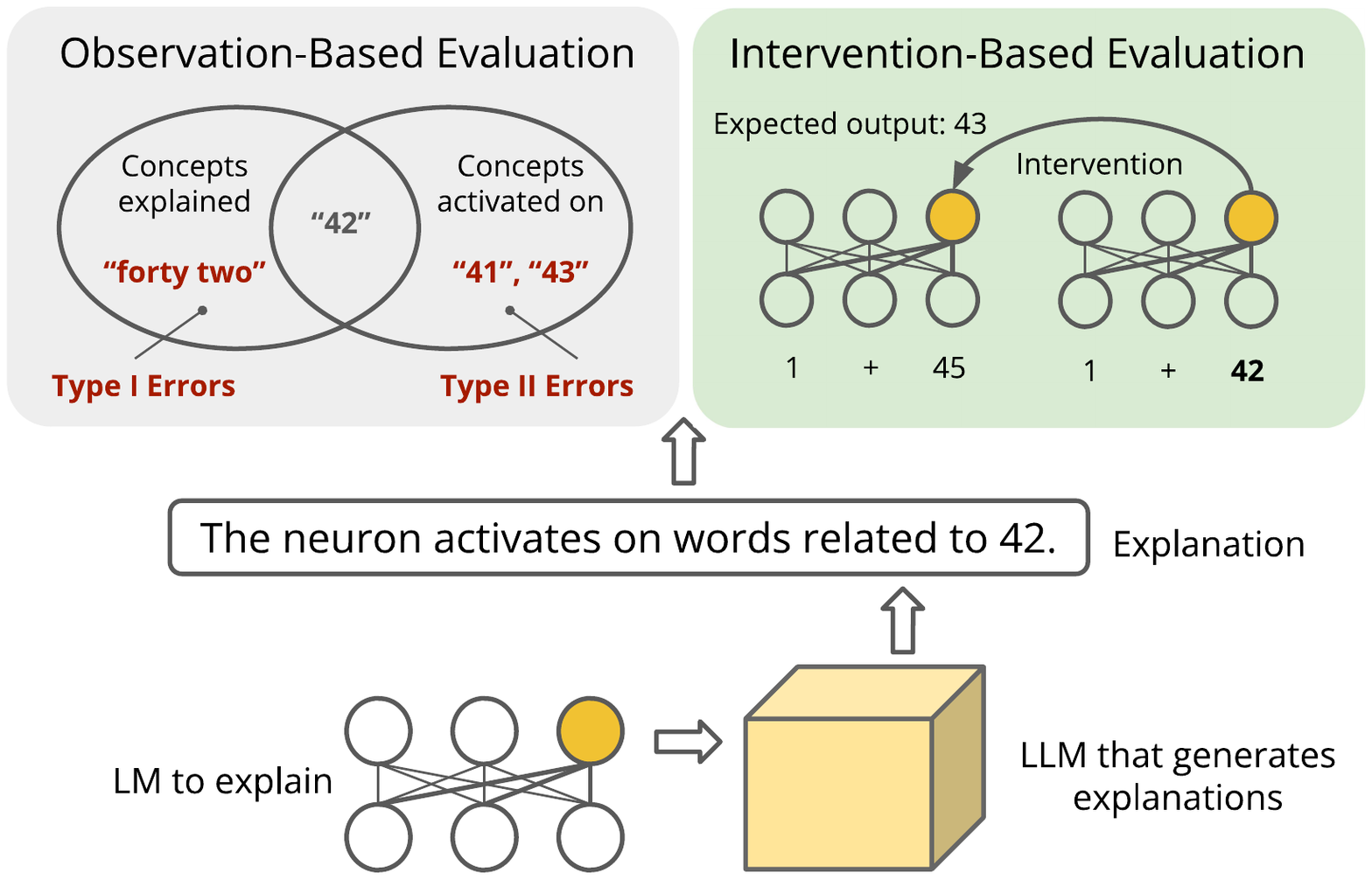}
     \vspace{-7ex}
     \caption{An overview of our proposed framework. In the \textit{observational mode}, we evaluate whether a neuron activates on strings picked out by the explanation. In the \textit{intervention mode}, we assess whether the neuron is a causal mediator of the concept in the explanation.}
     \label{fig:overview}
     \vspace{-3ex}
 \end{figure}

In the current paper, we seek to define criteria for assessing natural language explanations that claim individual neurons represent a concept in a text input. 
We consider two modes of evaluation (Figure~\ref{fig:overview}).
In the \emph{observational mode}, we evaluate the claim that a neuron $a$ activates on all and only input strings that refer to a concept picked out by the proposed explanation $E$.
Relative to a set of inputs, we can then use the error rates to assess the quality of $E$ for $a$. 

The observational mode only evaluates whether a concept is \textit{encoded}, as opposed to  \textit{used} \cite{antverg2022on}. Thus, we propose an \emph{intervention mode} to evaluate the claim that $a$ is a causally active representation of the concept denoted by $E$. We construct next token prediction tasks that hinge on the concept and intervene on the neuron~$a$ to study whether the neuron is a causal mediator of concepts picked out by $E$.

For example, consider the explanation \explanation{years between 2000 and 2003} of a neuron $a$.
In the observational mode, we experimentally test which strings the neuron $a$ activates on and quantify how closely this is aligned with the explanation's meaning.
In the intervention mode, we can construct a task where the prefix ``The year after $Y$ is'' is given and the model consistently outputs ``$Y+1$''. Then we can swap the value of $a$ for the value it takes on a different input and observe whether the behavior exhibits the expected change. The success rate of interventions quantifies the extent to which the neuron $a$ is a causal mediator of the concept of years \citep{vig2020causalmediation,geiger2021causal,Geiger-etal:2023:CA}.

To illustrate the value of this evaluation framework, we report on a detailed audit of the explanation method of \citet{bills2023language}, which uses GPT-4 to generate natural language explanations of neurons in a pretrained GPT-2~XL model. This is, at present, the largest-scale effort to automatically generate explanations of LLMs: the authors offer explanations for 300K neurons in GPT-2~XL. Automatically generating natural language explanations is inherently exciting, but our findings are inauspicious. In the observational mode, we find that even among the top 0.6\% of neurons which are considered well-explained by GPT-4's own assessment, the explanation is far from faithful; construed as predictions about neuron activations, GPT-4 generated explanations achieve a precision of 0.64 and a recall of 0.50.
In the intervention mode, the picture is more worrisome: we are unable to find evidence that neurons are causal mediators of the concepts denoted by the explanations. While the proposed explanations from the method of \citet{bills2023language} can be useful in exploring hypotheses about model computations, users of the method should have full knowledge of these assessments if they plan to make decisions based off these explanations.

We conclude by discussing some of the fundamental issues at hand. First, is natural language a good vehicle for model explanations? It seems appealingly accessible and expressive, but its ambiguity, vagueness, and context dependence are substantial problems if we want to use these explanations to guide technical decision making. Second, are neurons appropriate units to analyze? There may be useful signals in individual neurons, but it seems likely that the important structure will be stored in more abstract and distributed ways \cite{PDP1,PDP2,Smolensky1988, geva-etal-2022-transformer, Geiger-etal:2023:DAS}.

\section{Related Work}

\paragraph{Natural Language Explanations}

Explanations of black box AI models that come in the form of language text have the obvious benefit of being expressive and readable (\citealt{Hendricks2016VisualEx, Ling2017ExplainMath, Kim2018CarX, Do2020SNLIVE,  Kayser2022Xray}; see \citet{Wiegreffe2021reviewNatX} for a review). Recent work on automated neuron interpretability leverages natural language to produce neuron descriptions at scale \cite{hernandez2022natural,bills2023language,singh2023explaining}.

However, automated generation poses challenges for evaluation. The faithfulness of natural language explanation is inherently hard to evaluate \cite{atanasova-etal-2023-faithfulness}. Existing automated metrics are mostly neuron-level \cite{bills2023language,singh2023explaining}. Only a few measure model behaviors via ablation or editing \cite{hernandez2022natural}, which is critical for distinguishing \emph{encoded} vs.~\emph{used} information in neuron analysis \cite{antverg2022on}.

Besides concerns in faithfulness, recent work on distributed representations \cite{geva-etal-2022-transformer,Geiger-etal:2023:DAS} and superposition phenomena \cite{elhage2022superposition} suggests individual neurons may not provide the most interpretable structure.

\paragraph{Intervention-Based Methods} Interpretability methods that use interventions to create counterfactual model states have so far provided the most provably faithful explanations of model behaviors \cite{sundararajan2017axiomatic,chattopadhyay19a,vig2020causalmediation,feder-etal-2021-causalm,geiger2021causal,Geiger-etal:2023:CA,Geiger-etal:2023:DAS,meng2022locating,meng2023massediting,Materzynska2022DisentanglingVA,olsson2022context, wang2023interpretability,conmy2023automated}. Intervention-based methods are also adopted to measure the faithfulness of explanations \cite{antverg2022on,abraham2022cebab,atanasova-etal-2023-faithfulness}. Our evaluation is a causal mediation analysis \citep{PearlMediation, vig2020causalmediation}, a special case of causal abstraction analysis \citep{geiger2021causal,Geiger-etal:2023:CA}.
\section{Observation-Based Evaluation}\label{sec:obs}

We now define a framework for evaluating claims that a natural language text $E$ explains a neuron $a$ in a model $M$ using direct observational data.

\subsection{Methods}

We first need to specify how $E$ itself should be understood. Intuitively, an explanation like \explanation{years between 2000 and 2003} refers to a set of abstract entities (a specific set of years).\footnote{Does the English expression \word{between X and Y} include \word{X} and \word{Y}? The answer is highly variable and depends on the context and the entities being discussed \citep{Potts:Levy:2015}. Here we adopt an inclusive sense. This actually illustrates a core challenge of using natural language for model explanations: the explanations often need their own explanations.} However, this approach to meaning is hard to operationalize in terms of language models, which deal only with strings, so we opt to construe meanings as sets of strings. For example, the explanation \explanation{years between 2000 and 2003} of a neuron $a$ is given by $\sem{\text{years between 2000 and 2003}} = \{\text{``2000''}, \dots, \text{``2003''}, \text{``the year before 2002''}, \ldots\}$. 

Abstractly speaking, the above means that every explanation denotes an infinite set of strings: there will typically be large numbers of sensible ways of describing entities, and more generally, for any $q \in \sem{E}$, we will also have ``$q$ and True'' $\in E$, where ``True'' is a tautology of some sort. However, experimentally, we can approximate these sets with finite sets of strings. For example, we might approximate $\sem{\explanation{years between 2000 and 2003}}$ with just the set  $\{\text{``2000''}, \text{``2001''}, \text{``2002''}, \text{``2003''}\}$ for a partial but still robust test of $E$. In what follows, we assume that $\sem{E}$ is always approximated by a finite set; the precise membership of this set is an important experimental detail.

Bringing the above ideas together, we say that $\Explain_{M, Q}(a, E)$ is the claim that, for every input $q \in Q$ to model $M$ containing neuron $a$, the activation $a(q)>0$ iff $q\in \sem{E}$. Here, $Q$ is an experimental dataset defined to include our approximation of $\sem{E}$ as well as strings that will allow us to probe for cases where $E$ predicts no activation for the neuron but we do see activation.  For example, to test \explanation{years between 2000 and 2003}, we might use $Q = \{\text{``2000''}, \ldots, \text{``2003''}, \text{``pizza''}, \text{``\$5.75''}\}$.

In the observational mode, we evaluate whether the neuron $a$ activates on \emph{all and only} strings in $Q \cap \sem{E}$. We quantify this by considering an explanation $E$ as making predictions about whether the neuron $a$ will activate on a given input $q$. 
Type~I errors occur where the explanation $E$ falsely predicts that the neuron $a$ will activate on a string $q\in\sem{E}$. 
Type~II errors occur where the explanation $E$ falsely predicts that the neuron will not activate on a string $q\notin\sem{E}$. 
For the year example above, an error is of Type~I when $a$ does not activate on \text{``2001''} in an input, and of Type~II when $a$ does activate for a string like ``pizza'' in an input.

As there are usually neurons in each layer sharing semantically similar explanations, we can also evaluate how well an explanation $E$ predicts the activations of a set of neurons $[a_0, \dots, a_n]$, i.e., a claim that for every input $q \in Q$, $f([a_0(q), \dots, a_n(q)]) > 0$ iff $q\in \sem{E}$, where $f(\bm{x})=\bm{w}\cdot \bm{x} + b$ is a linear probe parameterized by $\bm{w}$ and $b$. For each explanation $E$, we first learn a probe $f$ that maximizes the mutual information between $\sem{E}$ and the activations \cite{belinkov-2022-probing} and then evaluate the claim with the learned probe. The claim of a single neuron $a$ activates on all and only strings in $\sem{E}$ can be viewed as a special case where $f$ is an identity function.

\subsection{Experimental Setup}\label{sec:obs-setup}

\paragraph{Explanations to Evaluate} We randomly sampled 300 (18\%) of the 1.7k neurons whose explanations have a score of at least 0.8. The score (referred to as the \emph{GPT-4 score} below) represents the correlation coefficient between GPT-4 simulated neuron activation and actual neuron activation over a set of inputs sampled from the GPT-2~XL training corpus. \citet{bills2023language} say that a GPT-4 explanation with a score higher than 0.8 means that ``according to GPT-4 [the explanation] accounts for most of the neuron's top-activating behavior''.

\paragraph{Dataset} For each neuron $a$ with explanation $E$, we construct two sets of test sentences. One set probes for Type~I errors by evaluating the claim ``$a$ activates on $q\in\sem{E}$'' with a set of sentences each containing a string $q\in \sem{E}$. We prompt GPT-3.5-turbo (referred as GPT-3.5 below) to sample a list of 20 words or phrases in $\sem{E}$ and embed each word or phrase into a sentence context. The other set probes for Type~II errors by evaluating the claim ``$a$ \emph{only} activates on $q\in\sem{E}$'' with a set of sentences each containing a string that the neuron $a$ activates on. We search for token sequences that the neuron $a$ activates on over a large corpus, record the sentence context of the token sequence, and prompt GPT-3.5 to determine whether the token sequence is in $\sem{E}$. When evaluating a set of neurons, we sample extra sentences to train the probe.

We manually verified the correctness of the generated datasets. We found over 95\% of the sentences to be valid. Most mistakes were on explanations that involve form-based properties like spelling, as GPT-3.5 does not have direct access to character information in each token \citep{kaushal-mahowald-2022-tokens,huang-etal-2023-inducing}. These cases, however, are easy to check programmatically. For form-based explanation $E$, we use a regex-based program to determine if a string belongs to $\sem{E}$. Wrongly selected negative entities can also occur due to vagueness of the explanation, i.e., the concepts are related following one interpretation but not another. We exclude incorrectly generated and ambiguous sentences from our test sets.  

\paragraph{Metrics}

For a given explanation $E$ of neuron $a$ and a set of inputs $Q$, we define precision and recall as follows. Let $a(q)$ be the activation of neuron $a$ on pattern $q$, and let $T_{Q}$ be the set of true positive instances in $Q$, i.e. $T_{Q}=\{q : q \in Q, q \in \sem{E} \text{ and } a(q) > 0\}$. Then:
\begin{equation*}
\text{Precision}(a, Q, E) = 
    \frac{|T_{Q}|}{|\{q : q \in Q, q \in \sem{E}\}|}
\end{equation*}
\begin{equation*}
\text{Recall}(a, Q, E) = \frac{|T_{Q}|}{|\{q : q \in Q , a(q) > 0 \}|}
\end{equation*}
We then compute F1-score as the harmonic mean of precision and recall. 
In the case where $q$ spans multiple tokens, we apply max pooling over all tokens. In the case where multiple neurons are evaluated, we use $f([a_0(q), \dots, a_n(q)])$ instead of $a(q)$, where $f$ is learned from a training set.

\paragraph{Baselines} We consider random pairings of neurons with GPT-4 explanations as baselines. For an explanation $E$, we randomly select $N$ neurons from a given layer and evaluate $E$ against the activations of the randomly selected neurons.

\subsection{Results}

\begin{table}[t]
  \centering
  \small
  \begin{tabular}[b]{l@{}c c c c c}
 \toprule
  & No Probe & \multicolumn{4}{c}{With Probe} \\ 
  & N=1 & N=1 & N=2 & N=4 & N=16 \\ 
 \midrule
Random & 0.00 & 0.29 & 0.44 & 0.54 & 0.69 \\
GPT-4 & 0.56 & 0.60 & 0.64 & 0.67 & 0.73 \\
 \bottomrule
\end{tabular}
 \caption{F1 scores measure how well randomly selected explanations and GPT-4 generated explanations predict neuron activations, averaged over 300 explanations with a GPT-4 score of at least 0.8. For each explanation to evaluate, we either randomly select $N$ neurons or select $N$ neurons whose explanations are semantically most similar to the given explanation.}
 \label{tab:probing_results}
 \vspace{-3ex}
\end{table}
\begin{table*}[t!]
  \small
  \centering
  \resizebox{\textwidth}{!}{%
  \begin{tabular}{p{0.16\linewidth}|p{0.28\linewidth}|p{0.28\linewidth}|p{0.28\linewidth}}
 \toprule
Explanation & True Positives & Type I Errors & Type II Errors \\ 
 \midrule

\makecell*[{{p{1\linewidth}}}]{days of the week}  &  \makecell*[{{p{1\linewidth}}}]{\colorbox{red!12}{\strut I} \colorbox{red!16}{\strut  have} \colorbox{red!12}{\strut  a} \colorbox{red!2}{\strut  music} \colorbox{red!14}{\strut \underline{ class}} \colorbox{red!1}{\strut \underline{ every}} \colorbox{green!100}{\strut \underline{ Wednesday}} \colorbox{red!8}{\strut \underline{ evening}}}  &  \makecell*[{{p{1\linewidth}}}]{\colorbox{red!11}{\strut \underline{Thursday}} \colorbox{green!100}{\strut  is} \colorbox{red!17}{\strut  usually} \colorbox{red!9}{\strut  reserved} \colorbox{red!14}{\strut  for} \colorbox{red!10}{\strut  grocery}}  &  \makecell*[{{p{1\linewidth}}}]{\colorbox{red!11}{\strut Philadelphia} \colorbox{green!100}{\strut \underline{ is}} \colorbox{red!6}{\strut  where} \colorbox{red!14}{\strut  the} \colorbox{red!0}{\strut  Declaration} \colorbox{red!0}{\strut  of} \colorbox{red!0}{\strut  Independence}} \\ \midrule

\makecell*[{{p{1\linewidth}}}]{years, specifically four-digit years}  &  \makecell*[{{p{1\linewidth}}}]{\colorbox{green!64}{\strut  Castro} \colorbox{red!2}{\strut  took} \colorbox{red!13}{\strut  power} \colorbox{red!2}{\strut  in} \colorbox{red!14}{\strut  Cuba} \colorbox{red!2}{\strut  in} \colorbox{green!100}{\strut \underline{ 1959}} \colorbox{red!2}{\strut .}}  &  \makecell*[{{p{1\linewidth}}}]{\colorbox{red!2}{\strut rated} \colorbox{red!1}{\strut  during} \colorbox{red!15}{\strut  re} \colorbox{red!4}{\strut -} \colorbox{red!9}{\strut entry} \colorbox{red!5}{\strut  in} \colorbox{red!14}{\strut \underline{ 2003}} \colorbox{red!2}{\strut .}}  &  \makecell*[{{p{1\linewidth}}}]{\colorbox{red!12}{\strut We} \colorbox{red!1}{\strut  need} \colorbox{red!7}{\strut  to} \colorbox{green!100}{\strut \underline{ rev}} \colorbox{red!1}{\strut \underline{amp}} \colorbox{red!2}{\strut  the} \colorbox{red!11}{\strut  website} \colorbox{red!4}{\strut  to} \colorbox{red!0}{\strut  attract} \colorbox{red!8}{\strut  more}} \\ \midrule

\makecell*[{{p{1\linewidth}}}]{the word "most" and words related to comparison}  &  \makecell*[{{p{1\linewidth}}}]{\colorbox{red!0}{\strut  lottery} \colorbox{red!0}{\strut  is} \colorbox{red!0}{\strut  a} \colorbox{red!11}{\strut  singular} \colorbox{red!1}{\strut  event} \colorbox{red!11}{\strut  for} \colorbox{green!100}{\strut \underline{ most}} \colorbox{red!2}{\strut  people} \colorbox{red!5}{\strut .}}  &  \makecell*[{{p{1\linewidth}}}]{\colorbox{red!11}{\strut She} \colorbox{red!0}{\strut  is} \colorbox{red!0}{\strut  the} \colorbox{red!1}{\strut \underline{ most}} \colorbox{red!1}{\strut  talented} \colorbox{red!0}{\strut  artist} \colorbox{red!3}{\strut  in} \colorbox{red!5}{\strut  the} \colorbox{red!0}{\strut  group}}  &  \makecell*[{{p{1\linewidth}}}]{\colorbox{red!6}{\strut Their} \colorbox{red!0}{\strut  hostility} \colorbox{green!100}{\strut \underline{ towards}} \colorbox{red!16}{\strut  each} \colorbox{red!15}{\strut  other} \colorbox{red!6}{\strut  was} \colorbox{red!14}{\strut  palpable} \colorbox{red!8}{\strut .}} \\ \midrule

\makecell*[{{p{1\linewidth}}}]{color-related words}  &  \makecell*[{{p{1\linewidth}}}]{\colorbox{red!13}{\strut  the} \colorbox{red!6}{\strut  sky} \colorbox{red!8}{\strut  in} \colorbox{red!14}{\strut  vibrant} \colorbox{red!16}{\strut  shades} \colorbox{red!8}{\strut  of} \colorbox{green!43}{\strut \underline{ violet}} \colorbox{red!12}{\strut  and} \colorbox{green!100}{\strut  pink} \colorbox{red!7}{\strut .}}  &  \makecell*[{{p{1\linewidth}}}]{\colorbox{red!16}{\strut  garden} \colorbox{red!15}{\strut  bloom} \colorbox{red!5}{\strut ed} \colorbox{red!10}{\strut  in} \colorbox{red!16}{\strut  shades} \colorbox{red!9}{\strut  of} \colorbox{red!12}{\strut \underline{ mag}} \colorbox{red!17}{\strut \underline{enta}} \colorbox{red!9}{\strut .}}  &  \makecell*[{{p{1\linewidth}}}]{\colorbox{red!11}{\strut  her} \colorbox{red!16}{\strut  lifelong} \colorbox{red!4}{\strut  dream} \colorbox{red!11}{\strut ,} \colorbox{red!9}{\strut  she} \colorbox{green!100}{\strut \underline{ opened}} \colorbox{red!8}{\strut  her} \colorbox{red!16}{\strut  own} \colorbox{red!0}{\strut  bakery}} \\ \midrule

\makecell*[{{p{1\linewidth}}}]{reflexive pronouns related to people or entities}  &  \makecell*[{{p{1\linewidth}}}]{\colorbox{green!3}{\strut They} \colorbox{red!0}{\strut  blamed} \colorbox{green!100}{\strut \underline{ themselves}} \colorbox{green!59}{\strut  for} \colorbox{red!0}{\strut  the} \colorbox{red!5}{\strut  failure} \colorbox{red!7}{\strut .}}  &  \makecell*[{{p{1\linewidth}}}]{\colorbox{green!2}{\strut She} \colorbox{red!0}{\strut  prepared} \colorbox{red!14}{\strut \underline{ herself}} \colorbox{green!1}{\strut  for} \colorbox{red!11}{\strut  the} \colorbox{red!15}{\strut  interview} \colorbox{red!10}{\strut .}}  &  \makecell*[{{p{1\linewidth}}}]{\colorbox{green!2}{\strut She} \colorbox{red!0}{\strut  gave} \colorbox{red!0}{\strut  the} \colorbox{red!4}{\strut \underline{ do}} \colorbox{red!3}{\strut \underline{ork}} \colorbox{green!100}{\strut \underline{nob}} \colorbox{red!0}{\strut  a} \colorbox{red!2}{\strut  twist} \colorbox{red!0}{\strut  and} \colorbox{red!1}{\strut  the} \colorbox{red!16}{\strut  door}} \\ \midrule

\makecell*[{{p{1\linewidth}}}]{proper names, specifically names related to mathematicians, scientists, and artists}  &  \makecell*[{{p{1\linewidth}}}]{\colorbox{red!12}{\strut \underline{E}} \colorbox{green!22}{\strut \underline{instein}} \colorbox{red!8}{\strut 's} \colorbox{red!15}{\strut  theory} \colorbox{red!16}{\strut  of} \colorbox{red!16}{\strut  relativity} \colorbox{red!14}{\strut  revolution}}  &  \makecell*[{{p{1\linewidth}}}]{\colorbox{red!3}{\strut Stephen} \colorbox{red!10}{\strut \underline{ Hawking}} \colorbox{red!10}{\strut  was} \colorbox{red!16}{\strut  a} \colorbox{red!13}{\strut  renowned} \colorbox{red!16}{\strut  physicist}}  &  \makecell*[{{p{1\linewidth}}}]{\colorbox{red!14}{\strut A} \colorbox{red!15}{\strut  software} \colorbox{green!100}{\strut \underline{ engineer}} \colorbox{red!14}{\strut  needs} \colorbox{red!14}{\strut  to} \colorbox{green!100}{\strut  compose} \colorbox{green!34}{\strut  lines} \colorbox{red!14}{\strut  of}} \\ \midrule

\makecell*[{{p{1\linewidth}}}]{technology-related words, specifically focusing on Linux and robots}  &  \makecell*[{{p{1\linewidth}}}]{\colorbox{red!5}{\strut \underline{R}} \colorbox{red!10}{\strut \underline{aspberry}} \colorbox{green!16}{\strut \underline{ Pi}} \colorbox{red!4}{\strut  is} \colorbox{red!3}{\strut  a} \colorbox{red!0}{\strut  small} \colorbox{red!2}{\strut ,} \colorbox{red!15}{\strut  versatile}}  &  \makecell*[{{p{1\linewidth}}}]{\colorbox{red!6}{\strut \underline{Ub}} \colorbox{red!16}{\strut \underline{untu}} \colorbox{red!4}{\strut  is} \colorbox{red!3}{\strut  a} \colorbox{red!9}{\strut  user} \colorbox{red!1}{\strut -} \colorbox{red!0}{\strut friendly}}  &  \makecell*[{{p{1\linewidth}}}]{\colorbox{red!3}{\strut He} \colorbox{green!100}{\strut \underline{ obtained}} \colorbox{red!10}{\strut  a} \colorbox{red!11}{\strut  restraining} \colorbox{red!9}{\strut  order} \colorbox{red!11}{\strut  to} \colorbox{red!14}{\strut  prevent}} \\ \midrule

\makecell*[{{p{1\linewidth}}}]{verbs related to movement or running out of something}  &  \makecell*[{{p{1\linewidth}}}]{\colorbox{red!3}{\strut He} \colorbox{red!0}{\strut  decided} \colorbox{red!3}{\strut  to} \colorbox{green!100}{\strut \underline{ run}} \colorbox{red!7}{\strut  to} \colorbox{red!16}{\strut  the} \colorbox{red!13}{\strut  store} \colorbox{red!15}{\strut  before} \colorbox{red!16}{\strut  it}}  & \makecell*[{{p{1\linewidth}}}]{\colorbox{red!2}{\strut The} \colorbox{red!15}{\strut  clever} \colorbox{red!16}{\strut  fox} \colorbox{red!2}{\strut  managed} \colorbox{red!1}{\strut  to} \colorbox{red!0}{\strut \underline{ evade}} \colorbox{red!0}{\strut  capture}}    &  \makecell*[{{p{1\linewidth}}}]{\colorbox{red!3}{\strut He} \colorbox{red!10}{\strut  loves} \colorbox{green!100}{\strut \underline{ ice}} \colorbox{green!25}{\strut  cream} \colorbox{red!15}{\strut ,} \colorbox{red!16}{\strut  but} \colorbox{red!6}{\strut  on} \colorbox{red!13}{\strut  the}} \\ 

 \bottomrule
 \end{tabular}
}
\caption{Examples of GPT-4 generated neuron descriptions with correct and error cases. The \underline{underlined} words and phrases are strings belonging to the set denoted by the explanation. The ground truth GPT-2~XL neuron activation is color-coded, with activated tokens highlighted in \colorbox{green!50}{green}. Some examples are truncated due to space constraints.}
 \label{table:false_examples}
\end{table*}

Results over 300 neuron explanations are shown in Table \ref{tab:probing_results}. For single neuron without probing, the GPT-4 explanations have a mean F1 score of 0.56 (with a precision of 0.64 and a recall of 0.50), whereas the random baseline has a F1 score of zero. With learned probes, the F1 score of GPT-4 explanations is 0.60.  The F1-score has a correlation coefficient of $-$0.1 with the GPT-4 score. With more neurons, F1 scores increase while the margin over the random baseline decreases, suggesting that most semantically relevant neurons have already been sampled. Examples of error cases are shown in Table~\ref{table:false_examples}, with analysis in Appendix~\ref{sec:appendix_false_negatives}.

\subsection{Discussion}

Our experimental results show that the \citealt{bills2023language} explanations are not well aligned with neuron activations; with an F1 score around 0.6 across 300 of the top-scoring explanations, it seems as though it would be risky to depend on these explanations for downstream tasks.

One might wonder how it can be that high GPT-4 scores do not lead to high precision/recall in our evaluation.
There is no inconsistency here, though, and indeed it is easy to show that a high GPT-4 score does not guarantee a faithful explanation. 

The GPT-4 score is computed on a set of 10 examples from the GPT-2~XL training corpus, 5 containing tokens with top activations and 5 randomly sampled. We now show that an unfaithful explanation with a precision of 0.50 can still have a perfect GPT-4 score with high probability. Consider an unfaithful explanation $E =  \explanation{year 2000 and 2001}$ of a neuron $a$ that only activates on ``2000''. When sampling the 10 examples from a corpus that has $n\%$ examples containing ``2001'', the probability of having at least one example containing ``2001'' (a Type~I error) is $1-(1-n\%)^5\approx 5n\%$. For any large corpus, $n\%$ could be extremely small due to a long tail distribution, which means the GPT-4 score is insensitive to Type~I errors. In contrast, our precision metric can capture Type~I errors by directly sampling different instances from $\sem{E}$, such that 50\% test examples should contain ``2001''. 

This example shows two things: (i) high correlation scores from GPT-4 simulations do not guarantee high-quality explanations, and (ii) our observational testing regime is more reliable, provided the chosen experimental datasets have the potential to diagnose both Type~I and Type~II errors.

\section{Intervention-Based Evaluation}

The goal of intervention-based evaluation is to assess the claim that a neuron $a$ is a causal mediator of the concept denoted by $E$. Intervention-based evaluation allows us to distinguish concepts that are \emph{used} vs.~\emph{encoded} in a model \cite{antverg2022on}, which is tightly connected to applications that require control and manipulation of the model, such as model editing. If we would like to use the explanation to inform us about where a concept is stored, we need explanations that pass the intervention-based assessment. Otherwise, modifying neurons associated with the explanation will have no effect on model behaviors.

\subsection{Methods}

To conduct these analyses, we first identify a task that takes any string $q \in \sem{E}$ as part of the input and has an output behavior that depends on $\sem{E}$. To ensure that we are assessing $E$ rather than the model's performance, the task should be one that the model solves perfectly.

For example, consider a task where a model $M$ receives the prompt ``The year after $Y$ is'' and is evaluated on whether the next token is $Y+1$. Here, a set of inputs $Q_{E, T}$ for explanation $E = \explanation{years}$ is a set of inputs based in a single template $T$ = ``The year after $Y$ is'' and differing only in the substring $Y$, where $Y$ could be any string in $\sem{E}$ plus strings not in $\sem{E}$ that can be used to fill the template $T$, such as ``college''. $Q_{E, T}$ depends only on $E$ and $T$. We say $M$ performs this task perfectly if $M$ gets every case in  $Q_{E, T}$ correct.

In the intervention mode, we assess whether the neuron $a$ is a causal mediator between the string encoding the year $Y$ and the predicted tokens encoding the year $Y+1$. To do this, we require just a few technical concepts from the literature on causal mediation and causal abstraction.

Let $M(x)$ be the entire state of the model $M$ when it receives input $x$. In other words, $M(x)$ sets all the input, internal, and output representations of the model via a standard forward pass. Let $\tau$ be a function that maps an entire model state to some output behavior. In our example, $\tau$ could be a function that first (i) maps $M(\text{``The year after $Y$ is''})$ to the next token predicted via greedy decoding and then (ii) classifies that token as being the desired $Y+1$ value or not.

We use $\GetVals(M(x), v)$ to specify the value stored at the position $v$ in $M(x)$, and we use $M_{v\gets \textbf{i}}(x)$ to specify the intervention in which $M$ processes $x$ but the value at $v$ is replaced with the constant value $\textbf{i}$.

An \emph{interchange intervention} is a nested use of $\GetVals$ and the intervention operation. For a source input $s$ and an activation $a_t$ of the neuron $a$ at the step $t$, we set $\textbf{z} = \GetVals(M(s), a_t)$. For a distinct base input $b$, we then process $M_{a_t\gets \textbf{z}}(b)$. In other words, we process $b$ with everything as usual, except that the value of $a_t$ is the one it has when the model processes $s$.

With the above definitions, we can say that  $\CausalExplain_{M,\tau, T}(a, E)$ is the claim that for all inputs $b, s \in Q_{E, T}$, we have
\begin{equation}
\label{eq:cm}
\tau(M_{a_t\gets \textbf{z}}(b)) = \tau(M(s))
\end{equation}
where $\textbf{z} = \GetVals(M(s), a_t)$ for some step $t$.

This can be viewed as a variant of causal meditation \cite{vig2020causalmediation}.
In intuitive terms: given the prompt ``The year after $Y$ is'', the model returns the next year. If $a$ is causally explained by ``years'', and assuming $M$ performs our task perfectly, when we process ``The year after 2023 is'' but with the value of $a$ set to what it has when we process ``The year after 2000 is'', then the model should output 2001. If it outputs 2024 or some other token, then $a$ evidently did not encode ``years'' in a way that is causally efficacious for our task.

Finding even one task that satisfies these criteria is  strong evidence for the explanation. If we can't find such tasks, it is also evidence against the explanation; we might always worry that there are some tasks that do satisfy the criteria, but every failed task will erode our confidence that the explanation has any force in explaining model behavior.

\subsection{Experimental Setup}

\paragraph{Explanations to Evaluate}
\newcommand{\highlight}[2]{%
  \setlength{\fboxsep}{1.0pt}\colorbox{#1}{\texttt{#2}}
}

\begin{table*}
[t!]
  \small
  \centering
  \vspace{5ex}
  \resizebox{\textwidth}{!}{%
  \begin{tabular}{p{0.20\linewidth} p{0.18\linewidth} p{0.62\linewidth}}
 \toprule
\makecell*[{{p{1\linewidth}}}]{Explanation $E$} & Task & Template $T$ with \highlight{cyan!20}{strings} in $\sem{E}$ and expected \highlight{green!20}{outputs} \\
\midrule
Numbers (13\%)  & \makecell*[{{p{1\linewidth}}}]{Unit conversion \\
 Numerical comparison} &  \makecell*[{{p{1\linewidth}}}]{The hiking trail stretches for \highlight{cyan!20}{2} miles (\highlight{green!20}{3.2} \\
 The war was in \highlight{cyan!20}{1935} and he was born in \highlight{cyan!20}{1937}, which was a few years \highlight{green!20}{after}} \\
Verbs (9\%) & Verb tense & They \highlight{cyan!20}{play piano} every day, so I believe yesterday they also \highlight{green!20}{played piano} \\
Time expressions (0.3\%) & Verb tense & They play piano every day, so I believe \highlight{cyan!20}{yesterday} they also \highlight{green!20}{played piano} \\
Locations (4\%) & Capital retrieval & The capital of \highlight{cyan!20}{Canada} is \highlight{green!20}{Ottawa} \\

 \bottomrule
 \end{tabular}
}
\caption{Examples of intervention-based evaluation tasks. }
 \label{table:intervention_tasks}
\end{table*}

The explanations of interest are associated with neurons in the Transformer MLP (feed-forward) layers, where concepts are represented in a highly distributed manner that require inter-layer and intra-layer aggregation to decode \citep{geva-etal-2021-transformer}. Hence, we consider evaluating both explanations of individual neurons and explanations of a set of semantically similar neurons. For example, explanations related to numbers, such as \explanation{numbers, particularly two-digit numbers} and \explanation{numerical values related to quantity} are evaluated as a single abstraction of the concept \explanation{number}. We identify a few common concepts that cover 80K (27\%) of explanations that correspond to neurons at various layers, as shown in Table~\ref{table:intervention_tasks}.

\paragraph{Evaluation Tasks}

We curate two tasks per concept that involve different manipulations of the concept. Example tasks are shown in Table~\ref{table:intervention_tasks}.  

Evaluating on different tasks is necessary, as two neurons with the same vague explanation may have different functionalities. For example, neurons in the first layer may activate to detect a number, while neurons in middle layers may activate to compare two numbers, even if the hypothesized explanation for both neurons is \explanation{numerical values}. Depending on the functionality, we apply interchange interventions either at the token positions corresponding to the string in $\sem{E}$ or at the last token position. We include evaluation details in Appendix~\ref{sec:intervention_eval_details}.

\paragraph{Metrics}

For a given explanation $E$ of a set of neurons $[a_0, \dots, a_n]$, a task $T$, a set of input pairs $Q_{E, T}$, we define interchange intervention accuracy (IIA) as the percentage of input pairs where the intervention output matches the expected output according to \eqref{eq:cm}. This IIA metric can be seen as a variant of the metric of \citet{pmlr-v162-geiger22a}.

As many explanation methods also predict a confidence score with an explanation, we can extend the IIA metric to IIA@K, where given a set of neurons and an explanation $E$, the IIA is computed with respect to the top $K$ percent of neurons with the highest confidence score of $E$ being the explanation. IIA@K also allows us to compare explanations generated by two methods. Given a fixed set of neurons, such as all neurons in a given MLP layer, each method produces a ranking of which neurons are most likely explained by $E$. We then systematically vary $K$ to compare IIA@K between the two methods.

\paragraph{Baselines} To better understand to what extent a set of neurons could affect model behaviors, we also consider two baselines: a random baseline randomly selecting $K\%$ of neurons, and a token-activation correlation baseline selecting the top $K\%$ of neurons with high activation over tokens that represent instances in $\sem{E}$ and low activation over other tokens in the test inputs.  The random baseline serves as a lower bound on the causal effects, while the token-activation correlation baseline is expected to have stronger causal effects. A causal explanation should at least select neurons with an IIA@K higher than the random baseline.

\begin{table*}
[t!]
  \small
  \centering
  \resizebox{\textwidth}{!}{%
  \begin{tabular}[t]{|p{0.3\linewidth}|p{0.3\linewidth}|p{0.3\linewidth}|}
 \hline

\makecell*[{{p{1\linewidth}}}]{\includegraphics[trim={0 0 0 58},clip,width=1.0\linewidth]{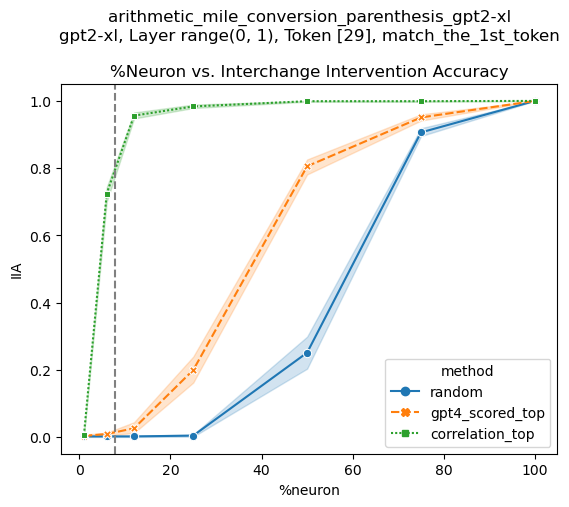} \\
\midrule
\textbf{Explanation}: Numbers \\
\textbf{Task}: Unit conversion \\
\textbf{Intervention location}: Layer 0 at the number tokens
} & 
\makecell*[{{p{1\linewidth}}}]{\includegraphics[trim={0 0 25 58},clip,width=1.0\linewidth]{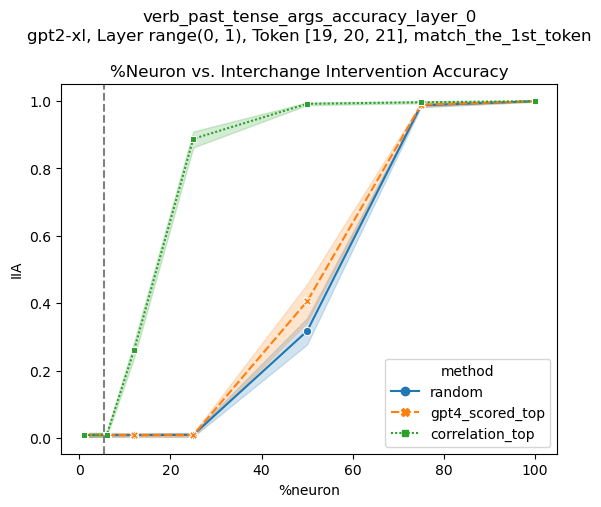} \\
\midrule
\textbf{Explanation}: Verbs \\
\textbf{Task}: Verb tense \\
\textbf{Intervention location}: Layer 0 at the verb tokens}& 
\makecell*[{{p{1\linewidth}}}]{\includegraphics[trim={0 0 0 58},clip,width=1.0\linewidth]{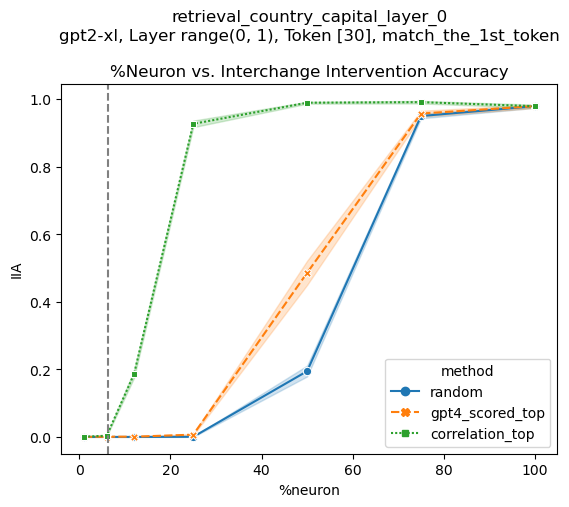} \\
\midrule
\textbf{Explanation}: Locations \\
\textbf{Task}: Capital retrieval \\
\textbf{Intervention location}: Layer 0 at the country tokens} \\
\hline
\makecell*[{{p{1\linewidth}}}]{\includegraphics[trim={0 0 0 58},clip,width=1.0\linewidth]{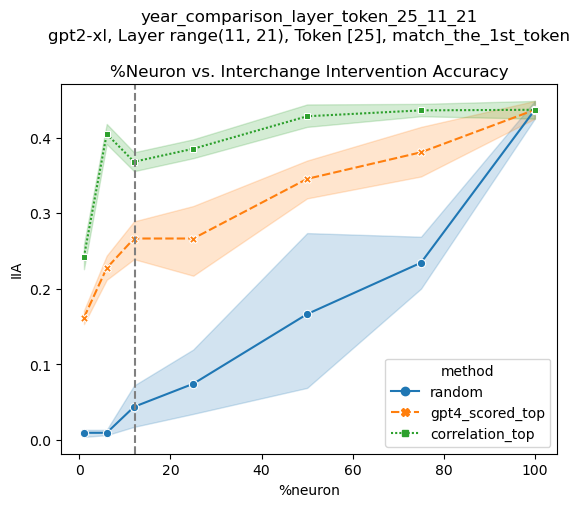} \\
\midrule
\textbf{Explanation}: Numbers (with years) \\
\textbf{Task}: Numerical comparison \\
\textbf{Intervention location}: Layer 11-21 at the second year tokens} &
\makecell*[{{p{1\linewidth}}}]{\includegraphics[trim={0 0 0 58},clip,width=1.0\linewidth]{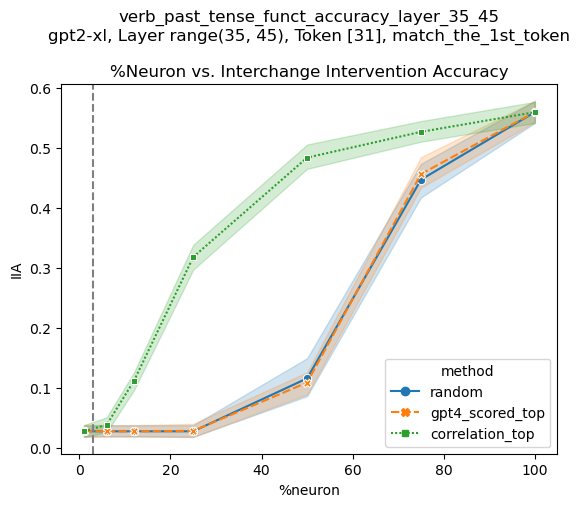} \\
\midrule
\textbf{Explanation}: Time expressions \\
\textbf{Task}: Verb tense \\
\textbf{Intervention location}: Layer 35-45 at the last token} &
\makecell*[{{p{1\linewidth}}}]{\includegraphics[trim={0 0 0 58},clip,width=1.0\linewidth]{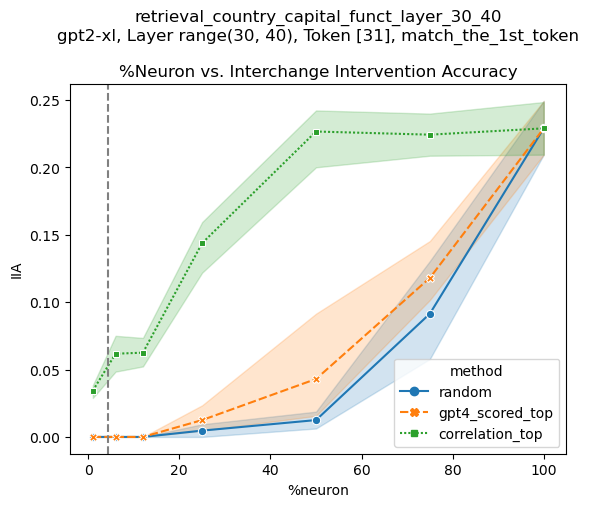} \\
\midrule
\textbf{Explanation}: Locations \\
\textbf{Task}: Capital retrieval\\
\textbf{Intervention location}: Layer 30-40 at the last token} \\

 \hline
 \end{tabular}
}
\caption{Intervention-based evaluation results. For each task, we rank and select the top K\% of neurons using three methods: random, correlation, and GPT-4 explanation score. We evaluate IIA@K for $K=1,6,12,25,50,75,100$. The dotted vertical line marks the percentage of GPT-4 explanation that directly mention the target pattern.}
\vspace{-5mm}
 \label{table:intervention_results}
\end{table*}

\subsection{Results}
\label{sec:inv_results}

Results on various tasks are shown in Table \ref{table:intervention_results}. There are two trends consistent across tasks. First, in terms of the IIA ranking, we have: token-activation correlation baseline $\gg$ GPT-4 explanation $\approx$ random baseline. Second, IIA increases as we intervene on a higher percentage of neurons. At $K=100$, MLP layer neurons show causal effects on all tasks. We further discuss the implications of these two observations below.

\subsection{Discussion}
\paragraph{Does GPT-4 produce causal explanations?} \mbox{GPT-4} generated explanations have similar causal effects as the random baseline on most tasks. The only exception is the explanation for neurons related to numerical expressions, which has higher IIA than the random baseline, but still far below the token-activation correlation baseline.

In other words, if we were using GPT-4 generated explanation to inform us which weights to modify in a model editing task, we would have similar performance as randomly selecting neurons to edit. This finding is worrisome but not surprising given low precision and recall values we obtained in our observational evaluation (Section~\ref{sec:obs}). 

\paragraph{Which neurons have causal effects?} The high IIA@100 suggests that MLP layer neurons, when evaluated as a whole, have strong causal effects on model behavior, especially in the first layer. Neurons in the middle and later layers only show causal effects on model behaviors after aggregating over multiple consecutive layers. This result is consistent with previous findings on the role of MLP layers \cite{geva-etal-2022-transformer,geva2023dissecting,meng2022locating}.

High IIA from the token-activation baseline suggests that the causal effects can be further narrowed down to neurons whose activation correlates well with the target pattern. For neurons in the first layer, the top 20\% of neurons with the highest correlation can already account for 80\% of the causal effect. While this finding shows there are relatively small subsets of neurons that encode certain high-level concepts, the granularity is still on the magnitude of hundreds of neurons. We have not found a task where intervening on a single neuron can change model behavior in a causal manner. We further discuss the choice for analysis unit in Section~\ref{sec:discussion-individual-neuron}.

\section{General Discussion}

\subsection{Inherent Drawbacks to Natural Language Explanations}

Is natural language the best medium for explaining large language models?

The benefits of using natural language in this context are that it is intuitive and expressive; one needn't learn a specialized formal language or data visualization language in order to consume explanations in this format and draw inferences from them to inform subsequent work.

However, natural languages are characterized by vagueness, ambiguity, and context dependence. These properties actually work in concert to facilitate the expressivity of language: vagueness and ambiguity allow words and phrases to be used flexibly, and context dependence means that people can coordinate on specific meanings using context \citep{Partee95}. From a relatively small set of primitives, we can talk about the complex universe we inhabit, but only because we can subtly refine the meanings of what we hear.

Given these facts about language, how are we meant to interpret explanations like the following, which were generated by the \citealt{bills2023language} method?
\begin{enumerate}[itemsep=0pt,topsep=6pt]
\item sentence-ending punctuation, specifically periods.
\item references to geographical locations, particularly related to Shanghai.
\item\label{exfinal} years, mostly from the 1980s and 2000s.
\end{enumerate}
Does the first explanation include the question mark, or does ``specifically periods'' refine the meaning to just the set containing the period? All of the above have the format ``a general concept $E$, specifically $E' \subset E$'', and there is no way to tell whether this is a prediction that the neuron will activate on $E \setminus E'$. Where the stakes are high, the human thing would be to discuss the meanings and the intentions behind them and come to some understanding. This path is not open to us for current LLM-based explanation methods, and it seems cumbersome if the goal is to use explanations to inform downstream tasks.

A similar issue arises where the explanation has the form  ``words and phrases related to a concept''. More than 30\% of neuron explanations in the \citealt{bills2023language} dataset contain the phrase ``related to''. Here are some examples: 
\begin{enumerate}[itemsep=0pt,topsep=6pt]
\item mentions of pizza and related food items 
\item words or parts of words related to the prefix `an'
\end{enumerate}
Is the first a reference to all Italian food, or to the various ingredients used to make pizza, or both? Is the second just a list of words beginning with those two characters, or does it refer to all words with one of the English morphological negations (e.g., ``an'', ``un'', ``in'', ``non'' and their allophones)?

There may be a way to define a fragment of natural language that is less prone to these interpretative issues, and then we could seek to have explainer models generate such language. However, if we do take these steps, we are conceding that model explanations actually require specialized training to interpret. In light of this, it may be better to chose an existing, rigorously interpreted formalism (e.g., a programming language) as the medium of explanation.

\subsection{Explanation Beyond Individual Neurons}
\label{sec:discussion-individual-neuron}

While top-activation patterns of individual neurons provide a rough idea of what concepts are encoded in the model, isolating the effect of individual neurons on model behavior is not always feasible, as features can be distributed across multiple neurons and may be polysemantic in nature \cite{antverg2022on,geva-etal-2022-transformer,elhage2022superposition, Geiger-etal:2023:DAS}. Our intervention-based evaluation results suggest that individual neurons are not the best unit of analysis in terms of understanding the causal effects of representations. 

Similarly, we should not limit ourselves to neurons located in particular parts of the network. While \citet{bills2023language} choose to analyze neurons in the MLP layers, attention heads and residual streams can also be used as different level of abstractions to understand model behaviors \cite{vig2020causalmediation,geiger2021causal,olsson2022context}.

\section{Conclusion}

We developed a framework for rigorously evaluating natural language explanations of neurons. Our \emph{observational mode} of analysis directly tests explanations against sets of relevant inputs, and our \emph{intervention mode} assesses whether explanations have causal efficacy. When we applied this framework to the method of \citet{bills2023language}, we saw low F1 scores in the observational mode and little or no evidence for causal effects in the intervention mode. Finally, we confronted what seem to us to be deep limitations of (i) using natural language to explain model behavior and (ii) focusing on neurons as the primary unit of analysis. Overall, we are more optimistic about approaches to model explanation that are grounded in structured formalisms (e.g., programming languages) and seek to explain how groups of neurons act in concert to represent examples and shape input--output behaviors. 

\section*{Limitations}

Our work contributes to improving the faithfulness of neuron interpretability methods that use natural language as a medium. Faithful explanation could provide the basis for safety assessments, bias detection efforts, model editing, and many other downstream applications. However, the ability to acquire more faithful explanations can also be used in malicious manipulations of the models. For example, high-quality explanations could help people to identify private or toxic information in a model, and these findings could be used to improve the model or to exploit the problem for ill-effect. We emphasize that explanations of large language models should always be used responsibly.

In an effort to evaluate the method proposed in \citet{bills2023language}, our analysis is primarily conducted on neuron behaviors of a pre-trained \mbox{GPT-2~XL} model, which is a decoder-only Transformer with 1.5B parameters \cite{radford2019gpt2}. The architecture used by \mbox{GPT-2~XL} has been widely adopted in current large language models, with similar neuron behaviors observed across variations of Transformers \cite{mu2020compositional,hernandez2022natural,geva-etal-2022-transformer,elhage2022superposition}, but we might nonetheless see different neuron behaviors emerge in new architectures. Our results should not be construed as extending directly to these architectures, but we are hopeful that our proposed evaluation framework will be useful for performing the necessary follow-up analyses.

\section*{Acknowledgements}

We thank William Saunders and Henk Tillman for helpful discussion of the evaluation framework. This research is supported in part by grants from Open Philanthropy, Meta, Amazon, and the Stanford Institute for Human-Centered Artificial Intelligence (HAI). 

\bibliography{anthology,custom}
\bibliographystyle{acl_natbib}

\appendix
\newpage
\clearpage
\onecolumn
\section*{Supplementary Materials}

\section{Prompts Used in Observation-Based Evaluation}
\label{sec:appendix}
Prompts used for test set construction are shown in Figure \ref{fig:prompt-selecting-positive} and Figure \ref{fig:prompt-gen-context}.

\begin{figure*}[hp]
    \centering
    \small
    \fbox{
    \begin{minipage}{0.95\textwidth}
    \texttt{\textcolor{gray}{PROMPT: For each word or symbol in ['los', 'middle', 'opened', 'green', 'phoenix', 'brown', 'grandchildren', 'north', 'silver', 'south', 'newly', 'long', 'narrow', 'big', 'white', 'small', 'hawaii', 'blue', 'central', 'san'], decide if it is related to the definition: color-related words. Output all the words you found related as a Python string list called related\_words. If no word is found, output related\_words = [].}}\\
    \texttt{\textbf{OUTPUT:} related\_words = ['green', 'brown', 'silver', 'white', 'blue']} \\
    \par
    \texttt{\textcolor{gray}{PROMPT: List 20 examples, each in 1-2 words, that are words expressing appreciation, importance, or overlooking. Directly output the examples as a Python string list called "examples".}}\\
    \texttt{\textbf{OUTPUT:} examples = ['grateful', 'thankful', 'admire', 'cherish', 'treasure', 'valuable', 'crucial', 'significant', 'vital', 'essential', 'crucial', 'imperative', 'paramount', 'indispensable', 'key', 'disregard', 'neglect', 'overlook', 'ignore', 'forgive']} \\
    \end{minipage}}
    
    \caption{Examples of prompts (gray) and GPT-3.5-turbo outputs (bold) for sampling string in $\sem{E}$.}
    \label{fig:prompt-selecting-positive}
\end{figure*}

\begin{figure*}[hp]
    \centering
    \small
    \fbox{
    \begin{minipage}{0.95\textwidth}
    \texttt{\textcolor{gray}{PROMPT: For each word in  ['grateful', 'thankful', 'admire', 'cherish', 'treasure', 'valuable', 'crucial', 'significant', 'vital', 'essential', 'crucial', 'imperative', 'paramount', 'indispensable', 'key', 'disregard', 'neglect', 'overlook', 'ignore', 'forgive'], write a coherent sentence that includes the word in a meaningful context and surround the text with two asterisks (**). Avoid having the text at the start of the sentence. Directly output the sentences as a python string list called "sentences".}}\\
    \texttt{\textbf{OUTPUT:} sentences = [ \\
    "I am **grateful** for all the support and love that I have received.",\\
    "I am **thankful** for having such wonderful friends in my life.",\\
    "I **admire** the way she handled that difficult situation with grace.",\\
    "I **cherish** the memories we made together during our vacation.",\\
    "**Treasure** the moments you spend with your loved ones.",\\
    "Education is **valuable** for personal and professional growth.",\\
    "A healthy diet is **crucial** for maintaining good physical health.",\\
    "**Significant** progress has been made in the field of technology.",\\
    "It is **vital** to take care of our mental well-being.",\\
    "Communication skills are **essential** for effective teamwork.",\\
    "A healthy work-life balance is **crucial** for overall well-being.",\\
    "It is **imperative** that we take action to protect the environment.",\\
    "Financial planning is **paramount** for a secure future.",\\
    "Innovation is **indispensable** in driving progress and growth.",\\
    "Time management is **key** to achieving success in any endeavor.",\\
    "We should not **disregard** the feelings of others.",\\
    "It is important not to **neglect** our physical health.",\\
    "Let's not **overlook** the efforts made by our teammates.",\\
    "We should not **ignore** the warning signs of burnout.",\\
    "**Forgive** yourself for any mistakes you have made in the past."
]}
    \end{minipage}}
    
    \caption{Examples of prompts (gray) and GPT-3.5-turbo outputs (bold) for embedding a string into a sentence context.}
    \label{fig:prompt-gen-context}
\end{figure*}

\section{Additional Analysis of Type I and Type II Errors}
\label{sec:appendix_false_negatives}

\begin{figure}[hp]
 \begin{subfigure}[b]{1\textwidth}
      \begin{minipage}{1\textwidth}
        \small
        \centering
        \resizebox{\textwidth}{!}{%
        \begin{tabular}{p{0.5\linewidth}|p{0.5\linewidth}}
        \toprule
GPT-4 explained patterns & Type II error patterns \\ 
\midrule
The pandemic had a \underline{negligible} impact on the economy. & The pandemic had a \underline{drastic} impact on the economy. \\
In her life, winning the lottery was a \underline{minor} turning point. & In her life, winning the lottery was a \underline{major} turning point. \\
The new regulations will have an \underline{insignificant} impact on businesses. & The new regulations will have a \underline{significant} impact on businesses. \\
In its research and development, the company made \underline{insubstantial} progress. & In its research and development, the company made \underline{substantial} progress. \\
To solve the problem, they introduced a \underline{conservative} new approach. & To solve the problem, they introduced a \underline{nonconservative} new approach. \\
The death of a loved one can have a \underline{superficial} effect on a person. & The death of a loved one can have a \underline{profound} effect on a person. \\
They received a \underline{paltry} amount of donations for the charity. & They received a \underline{considerable} amount of donations for the charity. \\
The dinosaur had a \underline{tiny} size compared to other animals. & The dinosaur had an \underline{enormous} size compared to other animals. \\
The desert stretched out before them, with its \underline{small} sandy dunes. & The desert stretched out before them, with its \underline{immense} sandy dunes. \\
She felt a \underline{mild} adrenaline rush before her performance. & She felt an \underline{intense} adrenaline rush before her performance. \\
The young artist's art exhibition received no recognition and was \underline{mediocre}. & The young artist's art exhibition received recognition and was \underline{noteworthy}. \\
Signing the peace treaty was a \underline{trivial} event in history. & Signing the peace treaty was a \underline{momentous} event in history. \\
The painting had \underline{unimpressive} color changes and simple details. & The painting had \underline{striking} color changes and intricate details. \\
The play had an \underline{unremarkable} plot twist that didn't surprise the audience. & The play had a \underline{dramatic} plot twist that surprised the audience. \\
His decision to invest in the company at an early stage was \underline{unimportant}. & His decision to invest in the company at an early stage was \underline{crucial}. \\
The news of the accident was \underline{inconsequential} and didn't affect the whole community. & The news of the accident was \underline{grave} and saddened the whole community. \\
The construction of a new airport was an \underline{modest} task for the engineers. & The construction of a new airport was a \underline{monumental} task for the engineers. \\
They had a \underline{light} discussion about the future of their relationship. & They had a \underline{serious} discussion about the future of their relationship. \\
         \bottomrule
        \end{tabular}}
      \end{minipage}
  \caption{Given the GPT-4 explanation ``small or minor changes, issues or improvements'', we generate minimal contrasting pairs where each adjective meaning minor is changed to its antonym. We extract neuron activation from each sentence at the \underline{underlined} words. If the GPT-4 explanation is accurate, the neuron should not activate on opposite words, however, we observe high activation on opposite words as shown in Figure \ref{fig:false_negative_example_distribution}.}
  \end{subfigure}
 \begin{subfigure}[b]{1\textwidth}
   \centering
  \includegraphics[trim={0 0 0 0},clip,width=0.45\linewidth]{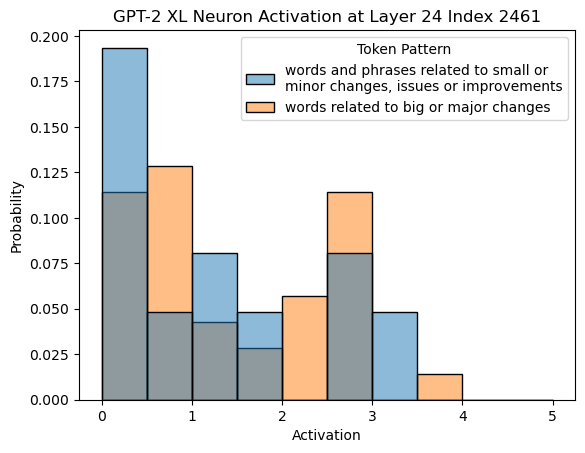}
  \caption{Neuron activation on ``big or major changes'' has similar distribution as ``small or minor changes'', despite GPT-4 explanation of the neuron is ``small or minor changes''.}
  \label{fig:false_negative_example_distribution}
  \end{subfigure}
  \caption{Examples of Type II errors where a neuron activates on antonyms of the concept in the explanation.}
  \label{fig:false_negative_antonyms}
\end{figure}

For Type I errors, i.e. precision error cases, we observe that form-based explanations have a higher precision at 0.78, while the rest only have a precision of 0.62.

For Type II errors, i.e. recall error cases, neurons mostly activate on inputs that have no clear relationship with the explained pattern, as hypothesized by work on superposition phenomena \cite{elhage2022superposition} where a single neuron potentially encodes a mix of concepts. We further investigate whether the Type~II errors in GPT-4 explanations are due to multiple concepts encoded in a single neuron, where the explanation only covers a subset of the concepts. 

We manually inspect 100 explanations that have Type II errors and observe at least 6 cases where the error involves antonyms of the concepts picked out by the explanation, such as the word ``above'' for an explanation \explanation{the word ``below'' and phrases related to it}, and the word ``ended'' for an explanation \explanation{words and phrases related to continuation, particularly in the context of `reading.'}. A full example with test inputs is shown in Figure \ref{fig:false_negative_antonyms}.

We also found neurons activate on inputs that have shared linguistic structures as the concepts in the explanation. For example, while the explanation is \explanation{days of the week}, the neuron also consistently activates on internet platforms such as ``Google'' or ``Facebook'' when preceded by the preposition ``on''. More interesting, the Type I errors of the same neuron involve inputs where the day of the week is not preceded by the preposition ``on''.

The majority of error cases, however, involve neurons activating on inputs unrelated to the explanation but nonetheless forming coherent concepts. These findings further support the view that individual neurons might not be the most useful unit of analysis in a large language model.

\section{Experiment Details in Intervention-Base Evaluation}
\label{sec:intervention_eval_details}

\begin{table*}
[t!]
  \small
  \centering
  \vspace{5ex}
  \resizebox{\textwidth}{!}{%
  \begin{tabular}{p{0.16\linewidth} p{0.19\linewidth} p{0.65\linewidth}}
 \toprule
\makecell*[{{p{1\linewidth}}}]{Explanation $E$} & Task & Template $T$ with \highlight{cyan!20}{strings} in $\sem{E}$ and expected \highlight{green!20}{outputs} \\
\midrule
\makecell*[{{p{1\linewidth}}}]{Numbers \\ (13\%)}  & \makecell*[{{p{1\linewidth}}}]{Unit conversion \\ Numerical comparison} &  \makecell*[{{p{1\linewidth}}}]{The hiking trail stretches for \highlight{cyan!20}{2} miles (\highlight{green!20}{3.2} \\
 The war was in \highlight{cyan!20}{1935} and he was born in \highlight{cyan!20}{1937}, which was a few years \highlight{green!20}{after}} \\
\makecell*[{{p{1\linewidth}}}]{Verbs \\ (9\%)} &  \makecell*[{{p{1\linewidth}}}]{Verb tense \\ Transitive/Intransitive} & \makecell*[{{p{1\linewidth}}}]{They \highlight{cyan!20}{play piano} every day, so I believe last night they also \highlight{green!20}{played}  \\ We live. They have pets. You leave. I stand. It happens. We \highlight{cyan!20}{swim}\highlight{green!20}{.\vphantom{placehoder}}}  \\
\makecell*[{{p{1\linewidth}}}]{Locations \\ (4\%)} & \makecell*[{{p{1\linewidth}}}]{Capital retrieval \\ City retrieval} & \makecell*[{{p{1\linewidth}}}]{The capital of \highlight{cyan!20}{Canada} is \highlight{green!20}{Ottawa} \\ 
The \highlight{cyan!20}{CN Tower} is located in the city of \highlight{green!20}{Toronto}} \\
\makecell*[{{p{1\linewidth}}}]{Names of people \\ (1\%)} &  \makecell*[{{p{1\linewidth}}}]{Gender agreement \\
Position retrieval} &  \makecell*[{{p{1\linewidth}}}]{\highlight{cyan!20}{Alice} didn't come because \highlight{green!20}{she} \\ \highlight{cyan!20}{Kay Ivey} was the governor of \highlight{green!20}{Alabama}
} \\
\makecell*[{{p{1\linewidth}}}]{Time expressions \\ (0.3\%)} & \makecell*[{{p{1\linewidth}}}]{Verb tense \\ Next day \\ } & \makecell*[{{p{1\linewidth}}}]{They play piano every day, so I believe \highlight{cyan!20}{last night} they also \highlight{green!20}{played} \\
Yesterday was \highlight{cyan!20}{Wednesday, February} 1st 2020. Today is \highlight{green!20}{Thursday}} \\
\makecell*[{{p{1\linewidth}}}]{Plural inflection \\ (0.1\%)} &  \makecell*[{{p{1\linewidth}}}]{Subject-verb agreement \\ Noun-pron. agreement} &  \makecell*[{{p{1\linewidth}}}]{We saw the \highlight{cyan!20}{trees}, which \highlight{green!20}{were} \\
The \highlight{cyan!20}{cats} ran away because \highlight{green!20}{they}} \\
 \bottomrule
 \end{tabular}
}
\caption{The full list of intervention-based evaluation tasks. }
 \label{table:intervention_tasks_all}
\end{table*}

\begin{table*}
[t!]
  \small
  \centering
  \resizebox{\textwidth}{!}{%
  \begin{tabular}[t]{|p{0.3\linewidth}|p{0.3\linewidth}|p{0.3\linewidth}|}
 \hline

\makecell*[{{p{1\linewidth}}}]{\includegraphics[trim={0 0 0 58},clip,width=1.0\linewidth]{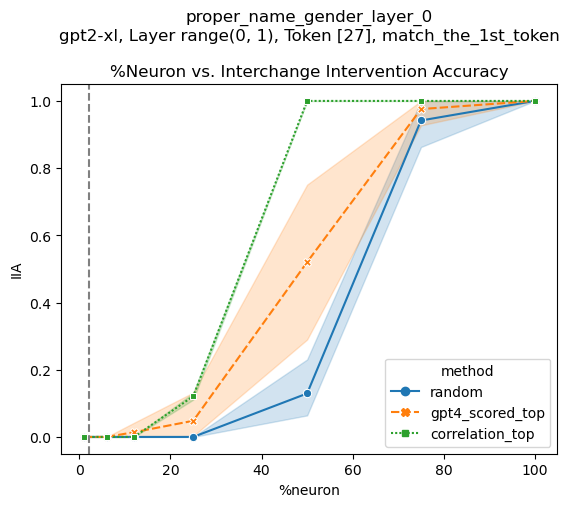} \\
\midrule
\textbf{Explanation}: Names of people \\
\textbf{Task}: Gender agreement \\
\textbf{Intervention location}: Layer 0 at the name tokens
} & 
\makecell*[{{p{1\linewidth}}}]{\includegraphics[trim={0 0 0 58},clip,width=1.0\linewidth]{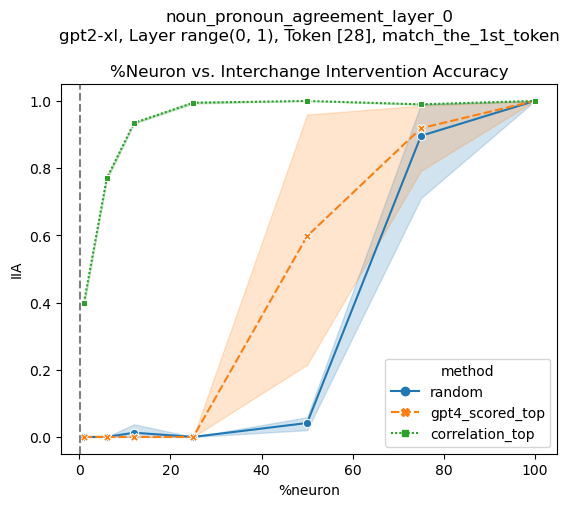} \\
\midrule
\textbf{Explanation}: Plural inflection \\
\textbf{Task}: Noun-pron. agreement \\
\textbf{Intervention location}: Layer 0 at the noun tokens}& 
\makecell*[{{p{1\linewidth}}}]{\includegraphics[trim={0 0 0 58},clip,width=1.0\linewidth]{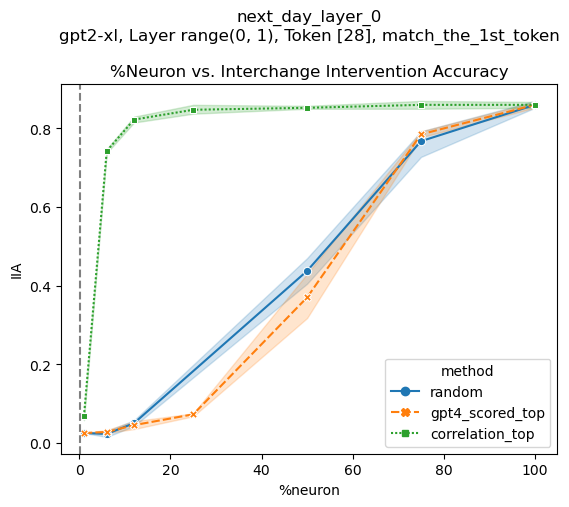} \\
\midrule
\textbf{Explanation}: Time expressions \\
\textbf{Task}: Next day \\
\textbf{Intervention location}: Layer 0 at the day of the week/month tokens} \\
\hline
\makecell*[{{p{1\linewidth}}}]{\includegraphics[trim={0 0 0 58},clip,width=1.0\linewidth]{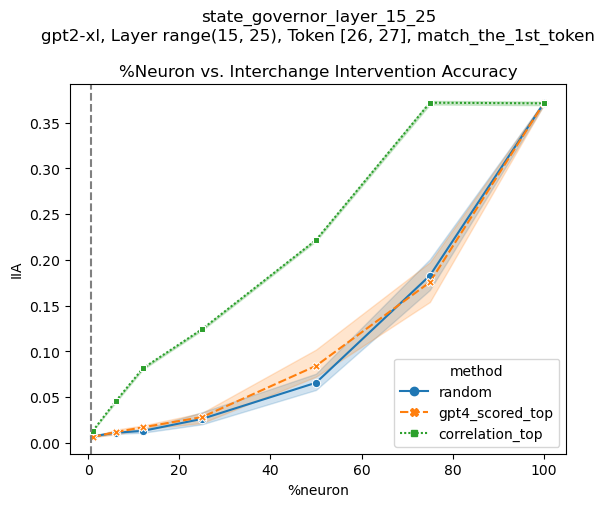} \\
\midrule
\textbf{Explanation}: Names of people \\
\textbf{Task}: Position retrieval \\
\textbf{Intervention location}: Layer 15-25 at the name token} &
\makecell*[{{p{1\linewidth}}}]{\includegraphics[trim={0 0 0 58},clip,width=1.0\linewidth]{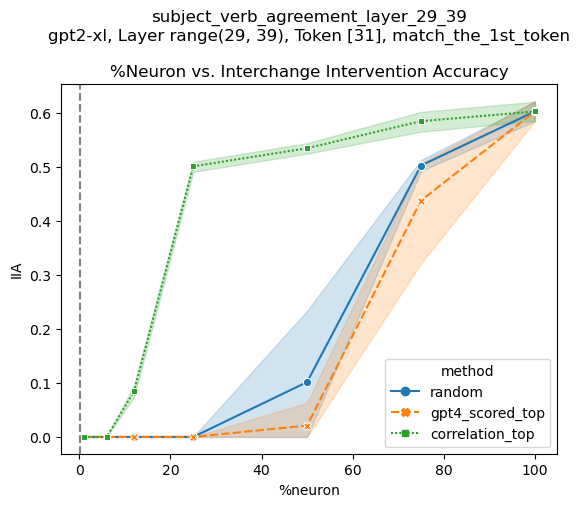} \\
\midrule
\textbf{Explanation}: Plural inflection \\
\textbf{Task}: Subject-verb agreement \\
\textbf{Intervention location}: Layer 29-39 at the last token} &
\makecell*[{{p{1\linewidth}}}]{\includegraphics[trim={0 0 0 58},clip,width=1.0\linewidth]{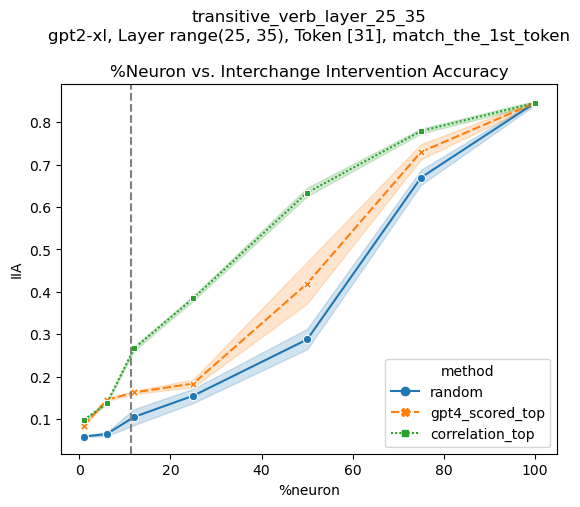} \\
\midrule
\textbf{Explanation}: Verbs \\
\textbf{Task}: Transitive/Intransitive\\
\textbf{Intervention location}: Layer 25-35 at the last token} \\

 \hline
 \end{tabular}
}
\caption{Additional intervention-based evaluation results.}
 \label{table:intervention_results_extra}
\end{table*}

\subsection{Tasks}
We curate tasks based on existing work that conducts behavioral testing on Transformer models, such as tests on grammatical phenomena \cite{warstadt-etal-2020-blimp-benchmark} and factual associations \cite{meng2022locating}. For each task specified by the template $T$ and a fixed set of at least 30 strings in $\sem{E}$, we verify that GPT-2~XL can correctly predict the next token on this set of inputs.
The full list of tasks is shown in Table~\ref{table:intervention_tasks_all}.

\subsection{Interchange Interventions}

\paragraph{Inputs}
For a given template $T$, we sample a set of at least 30 strings from $\sem{E}$ to fill the template and randomly pair up the filled templates to create 256 pairs of (base, source) as the test inputs. 

\paragraph{Intervention Locations} For each set of explanations to evaluate, one could perform an exhaustive search over every token position and report the highest IIA among all positions. However, based on how information is processed in Transformer MLP layers \cite{geva-etal-2022-transformer,meng2022locating,meng2023massediting,merullo2023language}, we could determine intervention locations as follows. If the neurons associated with the explanations are in the earlier layers (i.e. layer 1-24), we apply interchange interventions at the token positions that correspond to the string in $\sem{E}$, i.e. tokens highlighted in light blue in Table \ref{table:intervention_tasks_all}. If the neurons are in later layers, we apply interchange interventions at the last token position.

\subsection{Additional Results}

We show additional intervention-based evaluation results in Table~\ref{table:intervention_results_extra}. Results on the rest of the tasks can be found in Table~\ref{table:intervention_results}. These results further confirm the two trends discussed in Section~\ref{sec:inv_results}, namely (i) token-activation correlation baseline $\gg$ GPT-4 explanation $\approx$ random baseline and (ii) IIA increases as we intervene on a higher percentage of neurons.

\end{document}